\documentclass[letterpaper, 10 pt, conference]{ieeeconf}  % Comment this line out if you need a4paper

\IEEEoverridecommandlockouts                              % This command is only needed if 
\usepackage{graphicx}
\usepackage{epsfig} % for postscript graphics files
\usepackage{amsmath} % assumes amsmath package installed
\usepackage{amssymb}  % assumes amsmath package installed
\usepackage{amsbsy}  % assumes amsmath package installed
\usepackage{xcolor}
\definecolor{linkcolor}{RGB}{6,69,173}
\usepackage[
colorlinks=true,allcolors=linkcolor,pageanchor=true,plainpages=false,pdfpagelabels,bookmarks,bookmarksnumbered,backref=page
]{hyperref}

\usepackage{float}
\usepackage{capt-of}
\usepackage{balance}
\usepackage[nameinlink,capitalise]{cleveref}

\DeclareUnicodeCharacter{2212}{-}  % learning rate 1e-4

\crefname{figure}{Fig.}{Figs.}

\definecolor{ComColor}{RGB}{200,0,0}

\usepackage{xspace}
\newcommand{\ours}{NCF-v2\xspace}
\newcommand{\ncf}{NCF\xspace}
\newcommand{\ncforg}{NCF-v1\xspace}
\newcommand{\ncfmlp}{\ours\xspace{(MLP)}\xspace}
\newcommand{\ncftf}{\ours\xspace{(Transformer)}\xspace}
\newcommand{\mug}{mug-in-cupholder\xspace}
\newcommand{\bowl}{bowl-in-dishrack\xspace}

\newcommand{\ncfoutput}{\boldsymbol{p}\xspace}

\newcommand{\piprop}{$\pi_{prop}$\xspace}
\newcommand{\pitac}{$\pi_{tac}$\xspace}
\newcommand{\pincf}{$\pi_{\ours}$ (ours)\xspace}
\newcommand{\pigtc}{$\pi_{gtc}$ (oracle)\xspace}

\let\oldtwocolumn\twocolumn
\renewcommand\twocolumn[1][]{%
    \oldtwocolumn[{#1}{
    \begin{center}
        \vspace{-7mm}
        \includegraphics[width=\linewidth]{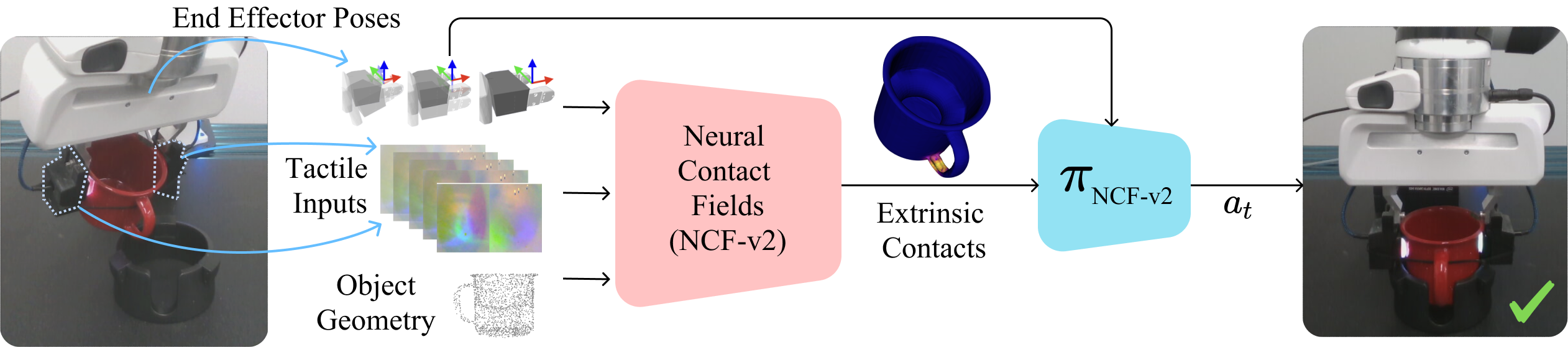}
        \vspace{-6mm}
        \captionof{figure}{Manipulation tasks such as inserting a mug in a cupholder involve interactions (extrinsic contacts) between the mug (object) and the cupholder (environment). Our model \ours enables sim-to-real transfer for estimating extrinsic contacts from tactile sensing between the robot fingers and the object. We use this representation to learn insertion policies that can be directly transferred to the real-world.}
       \label{fig:ncf_downstream_task}
       \vspace{-1mm}
    \end{center}
    }]
}

\title{\LARGE \bf
Perceiving Extrinsic Contacts from Touch\\Improves Learning Insertion Policies
}

\author{Carolina Higuera$^{1,2}$, Joseph Ortiz$^{1}$, Haozhi Qi$^{1,3}$, Luis Pineda$^{1}$,  Byron Boots$^{2}$, and Mustafa Mukadam$^{1}$\\[2mm]
$^{1}$FAIR, $^{2}$University of Washington, $^{3}$UC Berkeley
}

\begin{document}
\maketitle

\thispagestyle{empty}
\pagestyle{empty}

%%%%%%%%%%%%%%%%%%%%%%%%%%%%%%%%%%%%%%%%%%%%%%%%%%%%%%%%%%%%%%%%%%%%%%%%%%%%%%%%
\begin{abstract}
Robotic manipulation tasks such as object insertion typically involve interactions between object and environment, namely extrinsic contacts. 
Prior work on Neural Contact Fields (\ncf) use \emph{intrinsic} tactile sensing between gripper and object to estimate \emph{extrinsic} contacts in simulation. However, its effectiveness and utility in real-world tasks remains unknown.
In this work, we improve \ncf to enable sim-to-real transfer and use it to train policies for \mug and \bowl insertion tasks. We find our model \ours, is capable of estimating extrinsic contacts in the real-world. Furthermore, our insertion policy with \ours outperforms policies without it, achieving $33\%$ higher success and $1.36 \times$ faster execution on \mug, and $13\%$ higher success and $1.27 \times$ faster execution on \bowl.
% Project page: \url{https://github.com/carolinahiguera/NCF-Policy}
\end{abstract}

% 
%%%%%%%%%%%%%%%%%%%%%%%%%%%%%%%%%%%%%%%%%%%%%%%%%%%%%%%%%%%%%%%%%%%%%%%%%%%%%%%%
\vspace{-0.5mm}
\section{Introduction}
\vspace{-0.5mm}

Contact is inevitable; is estimation optional? During mundane activities like inserting a mug in a holder we often leverage our sense of touch to infer when and where the mug is in contact with the holder. We use this feedback, for example, when the mug's handle is stuck on top of the holder to guide it towards a position that allows for correct insertion. In this work, we study how robot manipulation policies can benefit from increased spatial awareness through estimation of extrinsic contacts.

Extrinsic contacts characterize the interactions between an object being manipulated and the environment. Such interactions are prevalent in contact-rich manipulation tasks, like peg insertion~\cite{dong2021tactile,schoettler2020deep,fu2022safely}, packing~\cite{dong2019tactile}, and tool manipulation (spatulas and wrenches)~\cite{wi2022virdo, sipos2022simultaneous}. Since ego-centric vision can be typically occluded or insufficient to capture subtle changes, modeling object-environment interactions would be challenging without heavy environmental and object instrumentation. Alternatively, tactile sensing can be more suitable to infer these interactions, albeit indirectly through sensing between robot hand and object.

Prior work on Neural Contact Fields (\ncf)~\cite{higuera2023neural} tackles the problem of estimating extrinsic contact between object and environment from vision-based tactile sensors on the gripper holding the object. It trains a neural field to estimate the probability of any 3D point on the object surface being in extrinsic contact without making any assumptions about the geometry of the contact. \ncf is able to localize multiple contact patches, captures contact/no-contact transitions, and generalizes to novel objects shapes in known categories. While these capabilities are a significant jump from related work at the time~\cite{prior_work_cline, prior_work_csp}, the results are limited to simulation and its applicability to downstream tasks is unexplored.

In this work, we analyze \ncf~\cite{higuera2023neural} in the real world and observe that the tactile embeddings in the model are prone to real-world sensory noises and the recurrent structure of the contact regressor leads to error accumulation. We make simple yet significant improvements to the tactile embedding and contact regressor, such that our model \ours can transfer to the real-world. Then, we use this representation to train policies in simulation for \mug and \bowl insertion tasks, and directly transfer them to the real-world (Fig.~\ref{fig:ncf_downstream_task}). To analyze the impact of estimating extrinsic contacts on policy learning, we comprehensively evaluate four policies with different observation inputs: only using robot's proprioception, with tactile data, with \ours, and ground truth extrinsic contacts. We find that observability over extrinsic contacts leads to policies with consistently higher success rates and greater task efficiency measured through number of time steps to success.

\vspace{-0.5mm}
\section{Related Work}
\vspace{-0.5mm}

Extensive work has explored localization and control of intrinsic contacts, i.e. interactions between end-effector and object, by focusing on internal forces of the grasp. Among those, it is common to use tactile sensing to obtain approximations of shear and normal forces of the grasp. Some works explore maintaining sticking contact~\cite{veiga2015stabilizing, li2014learning} or use sliding dynamics~\cite{dong2019maintaining, she2021cable}. Recent work has also proposed an interesting and simple approach to fuse proximity and visuo-tactile point clouds for contact patch segmentation between the sensing membrane and objects~\cite{yin2023proximity}. 

Research on extrinsic contacts is an emerging topic in the robotics community. Extrinsic contacts are prominent in contact-rich manipulation tasks, for example, insertion, use of tools, and in general tasks that require controlling the interactions of a grasped object with the environment. Ma et al.~\cite{prior_work_csp} discuss a theoretical basis for localizing environmental contacts, solving a constraint optimization problem for three types of contact geometry: point, line, and patch. Kim et al.~\cite{kim2023simultaneous} propose a method to simultaneously estimate and control extrinsic contact with tactile feedback. A factor graph is used to fuse a sequence of tactile and kinematic measurements to estimate and control the interaction between gripper-object-environment. 
% They also agree on the need to use sequences of tactile measurements over time to estimate the contact configuration.

For tool manipulation, Van der Merwe et al.~\cite{van2023learning} propose to learn a contact feature representation to predict the effects of the robot's actions on the contact between the tool and the environment, based on visuo-tactile sensing. This contact representation provides information about the binary contact state, the line of contact between the tool and the environment, and a predicted end-effector wrench. Along similar lines, Oller et al.~\cite{oller2023manipulation} show a side application of their method as a tracker of line contact between tool-environment, by querying points on the estimated object's SDF within a specific distance. Previous work on tracking extrinsic contacts has predominantly relied on touch sensing, utilizing either a force-torque sensor at the robot's wrist or vision-based tactile sensors. In contrast, Kim et al.~\cite{kim2023visionbased} propose a novel approach centered on learning to localize extrinsic contacts based solely on a single RGB-D camera view of the robot's workspace. The work most similar to ours is Neural Deforming Contact Field (NDCF)~\cite{van2023integrated}, where complex 3D contact patches are represented with neural fields. NDCF jointly models object deformations and contact patches via implicit representations, using a partial point cloud of the object and end-effector wrench information. This method allows tracking contacts on a deformable sponge in different environments.

% planning with tactile policies~\cite{dong2021tactile, okumura2022tactile, fu2022safely}.

%%%%%%%%%%%%%%%%%%%%%%%%%%%%%%%%%%%%%%%%%%%%%%%%%%%%%%%%%%%%%%%%%%%%%%%%%%%%%%%
\section{Sim-to-Real Neural Contact Fields}\label{sec:ncf}

% \hqi{should we call our NCF-v2? I currently feel section III.A mixes NCF-v1 and our improvement, so I made a few modifications. The original text is preserved in the comment.}

Our work builds on Neural Contact Fields (\ncf)~\cite{higuera2023neural} and improves it on several fronts to enable sim-to-real transfer. In this section, we first briefly discuss the original \ncf (Section~\ref{sec:ncfv1}) which we will call \ncforg for clarity. Then, we discuss the challenges and observations of transferring it to the real-world, and our subsequently improved model (Section~\ref{sec:ncfv2}) that we call \ours.

\subsection{\ncf preliminaries (\ncforg)}
\label{sec:ncfv1}

Neural Contact Fields~\cite{higuera2023neural} estimate the extrinsic contacts between surface points of an object and the environment. Given a set of $n$ query 3D points of the object, it outputs the probability of contact for each of them. 

\ncforg takes a shape representation, a history of end-effector poses, and a sequence of images from a vision-based touch sensor (e.g. DIGIT~\cite{digit_paper} is used in~\cite{higuera2023neural}) on the gripper as the inputs. Denoting $\ncfoutput_t \in \mathbb{R}^n$ as the extrinsic contact probabilities where $n$ is the number of query points, $\boldsymbol{g}_{t}$ as the embedding of a sequence of tactile observations, $\boldsymbol{r}_t$ as the object shape representation, $\boldsymbol{e}_t$ as a sequence of end-effector poses. Then \ncforg is a function $f$ with the following form: 
\begin{equation}\label{eq:ncf}
\ncfoutput_t = f_\theta(\ncfoutput_{t-1}, \boldsymbol{g}_{t-1}, \boldsymbol{r}_{t-1}, \boldsymbol{e}_{t-1}).
\end{equation}
It does not model the environment and only estimates local interactions between object and environments, and can generalize to unseen environments and shapes. However, the results in~\cite{higuera2023neural} are limited to simulation.

% When using a real DIGIT sensor and subtracting the background image, we retain information about color gradients in the contact patch. These gradients result from the motion caused by contact, which leads to surface displacements in the sensor's elastomer. To make the autoencoder robust to these variations, we find it helpful to train it using multiple non-contact images. In total, we employed 24 background images from real DIGIT sensors

% In this work, we study the utility of estimating extrinsic contacts, for downstream tasks. Neural Contact Fields (\ncf) can track such contacts in simulation, using vision-based tactile sensing between gripper and object. We build upon prior work to facilitate the sim-to-real transfer of \ncf and its posterior use for insertion tasks. For clarity, we will refer to the improved pipeline as \ncf and to prior work as \ncforg.

% \subsection{\ncf pipeline}

% \ncf estimates the probability of extrinsic contact for any 3D point on the object's surface, utilizing a sequence of tactile images from DIGIT sensors~\cite{digit_paper} located on the gripper's fingers, along with the most recent history of the robot's end-effector pose.  \ncf can localize multiple contact patches without making any assumptions about its geometry and it can capture contact/no-contact transitions, generalizing to complex object shapes (mugs, bottles, and bowls) and environment configurations. In Fig.~\ref{fig:neural_contact_field_arq} we show the improved \ncf pipeline and next, we describe each submodule and the modifications that we made to enable sim-to-real.

\begin{figure}[!t]
    \centering
    \vspace{2.5mm}
    \includegraphics[width=\linewidth]{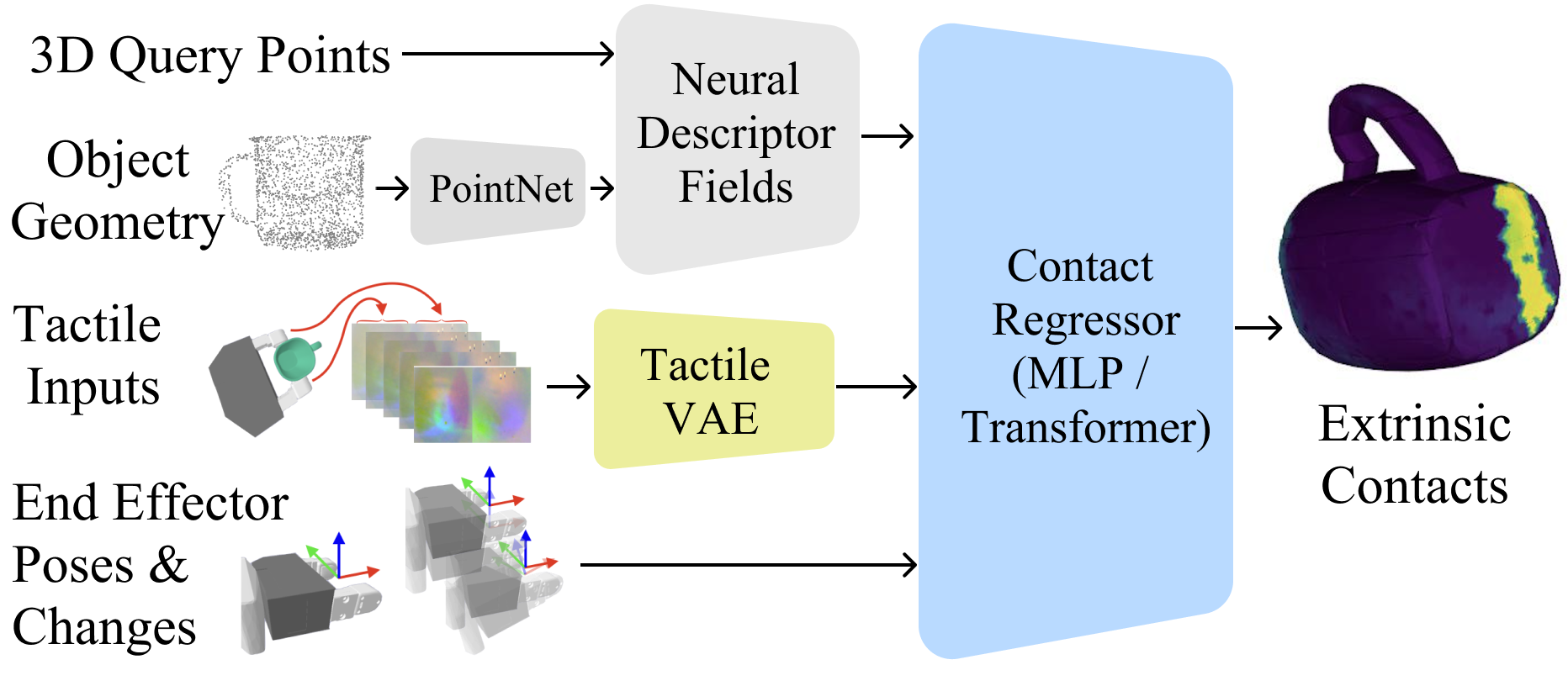}
    \vspace{-6.5mm}
    \caption{Neural Contact Fields v2 (\ours) architecture for tracking extrinsic contact (between object and environment) given tactile sensing (between robot hand and object). For a set of query 3D points on the object surface, \ours estimates the probability of any point in extrinsic contact.}
    \label{fig:neural_contact_field_arq}
    \vspace{-6mm}
\end{figure}

\begin{figure*}[!t]
    \centering
    \vspace{0.2mm}
    \includegraphics[width=\textwidth]{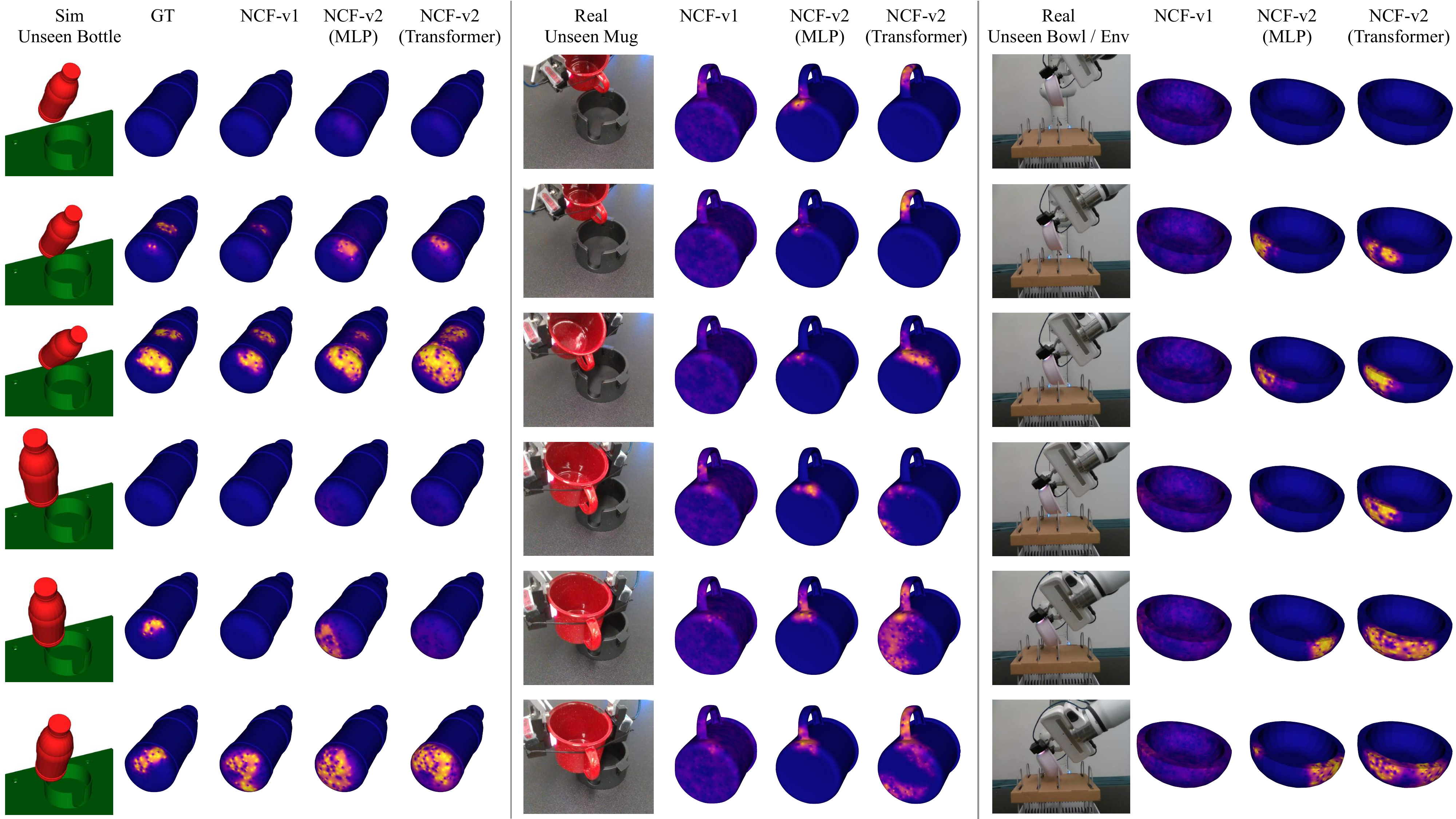}
    \vspace{-6.2mm}
    \caption{\textbf{Qualitative study with unseen bottle (left), mug (middle), and bowl in unseen environment (right) in simulation (left) and real-world (middle and right)}. The robot execute a predefined trajectory to make the object collide with the environment. We show estimated extrinsic contact patch from \ncforg, \ncfmlp, and \ncftf on the object surface. In simulation, all models exhibit comparable performance in estimating extrinsic contact on unseen object shapes, accurately tracking contact/no-contact transitions, and localizing multiple contact patches. In the real world, \ncftf shows best sim-to-real performance in estimating extrinsic contacts.}
    \label{fig:ncf_results}
    \vspace{-6mm}
\end{figure*}

\subsection{\ncf improvements (\ours)}
\label{sec:ncfv2}

In our experiments, we find it challenging to directly transfer \ncforg to the real-world. We observe two failure modes: 1) the tactile embedding $\boldsymbol{g}_t$ is not accurate enough since real-world tactile information is more noisy; 2) the recurrent structure in the contact regressor leads to error accumulation. To address these problems, we make the following improvements. The final architecture for Neural Contact Fields is presented in Fig.~\ref{fig:neural_contact_field_arq}.

% \vspace{0.05em}
\textbf{Tactile Embedding.} \ncforg trains an autoencoder to obtain tactile image embeddings at each timestep. A history of such embeddings is passed through a long short-term memory network (LSTM~\cite{hochreiter1997long}) to derive the final representations $\boldsymbol{g}_t$ for the sequence. We replace the autoencoder with a variational autoencoder~\cite{vae} to learn a regularized embedding space and empirically find it beneficial for sim-to-real transfer. We also pre-process each image by subtracting the corresponding background image before encoding. This modification allows the network to prioritize reconstructing the contact rather than reconstructing background RGB values. To make the autoencoder robust to variations across different touch sensors, we also randomize over different background images. In total, we employed 24 background images from real DIGIT sensors. After we get the embedding for each tactile image, we concatenate $T$ most recent embeddings and pass this vector through a Multi-layer Perceptron (MLP) instead of LSTM used in \ncf. We choose $T=5$ in our experiment and our sensor is sampled at 30 fps, which leads to $\sim$0.17 seconds of tactile signal history.

% \vspace{0.05em}
\textbf{Contact Regressor.} \ncforg uses a recurrent structure to estimate the extrinsic contact probability. However, we find this design makes sim-to-real transfer particularly challenging, as the errors tend to accumulate over time, especially during long executions. Inspired by the recent advances in temporal sensory fusion methods~\cite{qi2023general}, we remove this recurrent structure and only rely on the temporal cues from the inputs. Specifically, we remove $\boldsymbol{p}_{t-1}$ from Eq.~\eqref{eq:ncf}, and directly concatenate the other inputs to form a feature vector for the current timestep. After that, we feed this vector to a temporal aggregation network to get the contact probability of each query point. We experiment with two types of temporal aggregation networks: MLP and transformer.
The MLP version is a network with three layers, with output dimension [512, 128, 1]. Batch normalization and ReLU are appended after the first two hidden units. The transformer version is an encoder-only structure of two hidden layers with two heads. The feature dimension of transformer is 512 and the output is fed to two linear layers with output dimension [128, 1]. For both versions, a sigmoid activation in the end estimates the contact probability.

% \vspace{0.05em}
\textbf{Training.} We collect contact interactions using IsaacGym~\cite{makoviychuk2021isaac} and TACTO~\cite{tacto_sim}. We simulate a Franka Panda arm equipped with a parallel gripper and DIGIT sensors on the gripper. The robot executes trajectories in a cupholder scene (Fig.~\ref{fig:ncf_results}, left) with three object categories collected from 3D Warehouse~\cite{shapes_dataset}: mugs, bottles, and bowls. We curated a dataset containing five different shapes for each category and reserved one shape for testing shape generalization. All of the objects are initialized in a stable grasp pose. For neural descriptor field (NDF)~\cite{ndf}, we follow the same setting as~\cite{higuera2023neural} and use a pretrained NDF to obtain a feature vector for each query point. We separately trained the DIGIT VAE using tactile images from the TACTO~\cite{tacto_sim} simulator. See~\cite{higuera2023neural} for further details on prior architecture and training.

\section{\ours Evaluation}

% In this section, we study two questions: 1) Can \ours learned entirely in simulation used to estimate extrinsic contact in the real-world; 2) Can \ours representation enable efficient sim-to-real transfer on downstream tasks?

We first quantitatively compare \ncforg, \ncfmlp, \ncftf in simulation. We execute a predefined trajectory shown in Fig.~\ref{fig:ncf_results} (left) to make the object collide with the environment (a cupholder in this experiment). All three models can successfully track contact/no-contact transitions and localize multiple contact patches with complex geometries and demonstrate similar performance in estimating the contact patch on the unseen bottle. We compute the Mean Squared Error (MSE) over 100 test trajectories involving unseen mugs, bottles, and bowls. The MSE for \ncforg is 0.048, \ncfmlp is 0.042, and \ncftf is 0.039.
% , respectively. 
The results for \ncforg are consistent with~\cite{higuera2023neural} in simulation.

%However, when we try to deploy the trained model in the real-world, we observe \ncforg and \ours are significantly different. As shown in Fig.~\ref{fig:ncf_results} (middle and right), the \ncftf model demonstrates superior performance in capturing the contacts occurring between the mug and cupholder during experiments. It effectively identifies the contact points on the handle and smoothly transitions to detecting contact on the foot of the mug when it is positioned on the edge of the cupholder. On the other hand, \ncfmlp captures the contact on the handle but struggles with tracking the transition. Meanwhile, \ncforg fails to transfer to real-world data, mostly predicting zero contact everywhere. The predicted contact probabilities exhibit noise, and overall, the model faces difficulties in accurately localizing contacts.

% Regarding the test trajectory involving the unseen bowl, both \ncfmlp and \ncftf effectively capture the transition from no-contact to contact. However, \ncftf demonstrates superior localization of the contact patches. For instance, it accurately identifies that the bowl initially makes contact on the edge near the slot in the cupholder, and then slides backward until two sides of the bowl (along the z-axis of symmetry) are in contact after it is placed in the middle of the cupholder.

However, when we deploy the trained models in the real-world, we observe the performance of \ncforg is significantly worse than \ours. Fig.~\ref{fig:ncf_results} (middle and right) shows the \ncftf model demonstrates superior performance in capturing the contacts occurring between object and environment. In the case of the unseen mug (middle), \ncftf effectively identifies the contact points on the handle and smoothly transitions to detecting contact on the foot of the mug when it is positioned on the edge of the cupholder. \ncfmlp captures the contact on the handle but struggles with tracking the transition. On the other hand, \ncforg fails to transfer to real-world data, mostly predicting zero contact everywhere. The predicted contact probabilities exhibit noise and the model faces difficulties in accurately localizing contacts.

We also evaluate generalization for unseen shape \emph{and} environment (Fig.~\ref{fig:ncf_results}, right). \ncfmlp and \ncftf effectively capture the transition from no-contact to contact. Both models capture the contacts as the bowl descends between the dividers, although \ncftf is more consistent, particularly towards the end of the trajectory when the bowl contacts the rack base while still situated between the dividers. As \ncftf consistently outperforms other variants in estimating extrinsic contacts on real data, we use it in the following experiments in learning policies for insertion tasks.

\section{Downstream Tasks Using \ours}

We explore the effectiveness of estimating extrinsic contacts in downstream tasks of \mug and \bowl, which involve contact-rich interactions between object and environment. 

\subsection{Experimental setup}

\begin{figure*}[!t]
    \centering
    \vspace{0.2mm}
    \includegraphics[width=\textwidth]{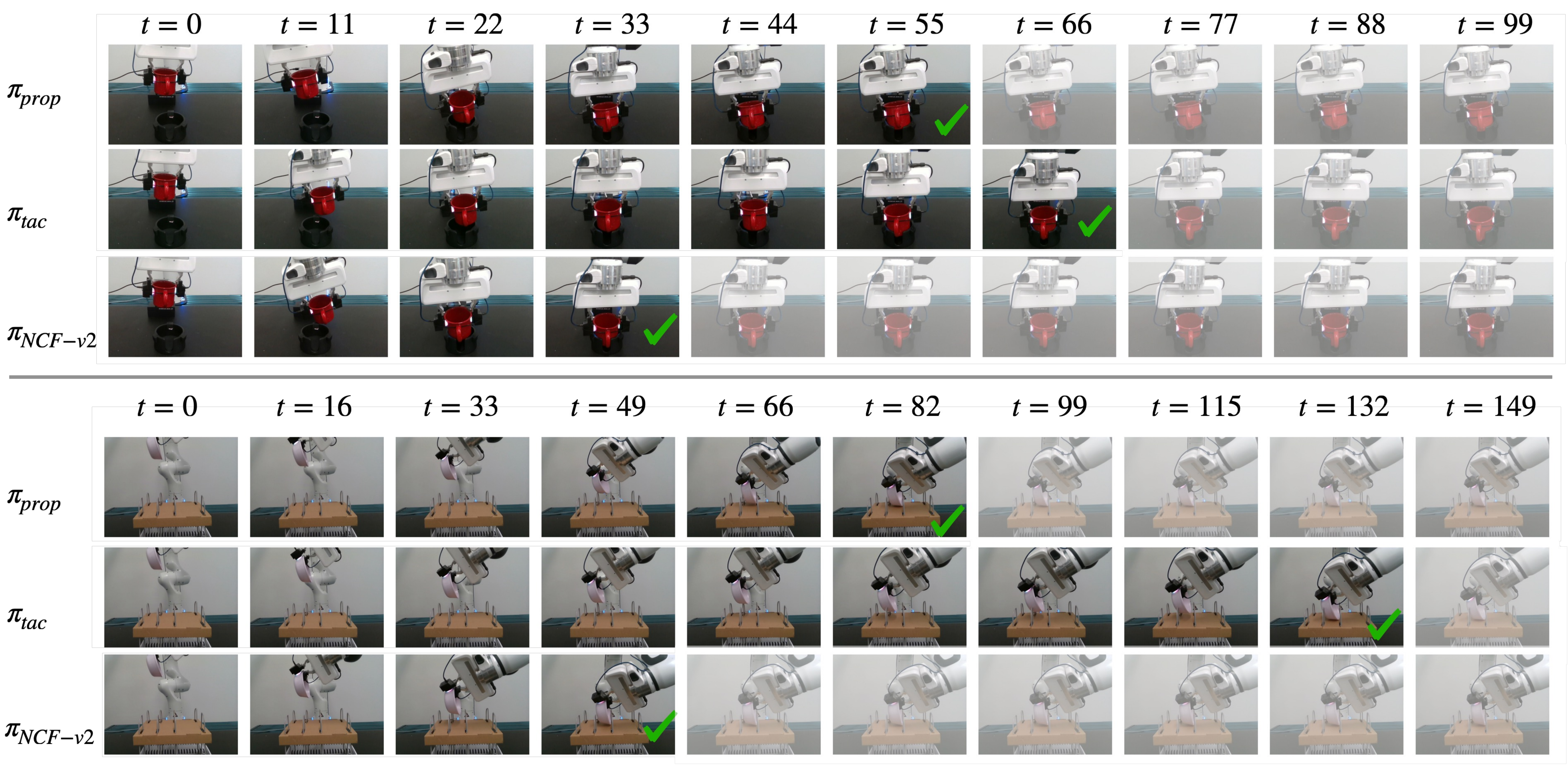}
    \vspace{-7.2mm}
    % \caption{Snapshots of successful and unsuccessful policy executions on \mug insertion task in real-world. On average, \pincf is $33\%$ more successful and $1.36 \times$ faster at completing the task, compared to other policies that do not leverage extrinsic contact.}
    \caption{Successful policy executions on \mug (top) and \bowl (bottom) insertion tasks in real-world. On average, \pincf is $13-33\%$ more successful and $1.27-1.36 \times$ faster at completing the tasks, compared to policies that do not leverage extrinsic contact.}
    \label{fig:rl_real_seq}
    \vspace{-3mm}
\end{figure*}

\textbf{Task 1: \mug.} A Franka Panda arm with a parallel gripper and DIGIT sensors on each finger starts with a mug grasped above a table. The goal is to insert the mug in a cupholder on the table. The cupholder features a slot on the side, as depicted in Fig.~\ref{fig:rl_real_seq} (top). Successful completion requires precise alignment ensuring the mug's handle fits into the cupholder's slot.

\textbf{Task 2: \bowl.} The robot starts with the bowl held above the dishrack. The goal is to insert the bowl into the second slot from the left, as depicted in Fig.~\ref{fig:rl_real_seq} (bottom). Successful completion requires approaching the target slot while maintaining awareness of the dividers.

For both tasks, the action space of the robot consists of 6-DOF transformations relative to the current end-effector pose, with the rotation expressed as axis-angle. These actions are set as the target for the joint-space IK controller. We use IsaacGym for simulating the tasks, using the Factory environments~\cite{narang2022factory} as reference, and use TACTO for simulating the touch sensors. For training, we run 256 environments in parallel, introducing randomization to the end-effector pose and running each episode for 250 steps.

\begin{figure}[!t]
    \centering
    \vspace{2.5mm}
    \includegraphics[width=\linewidth]{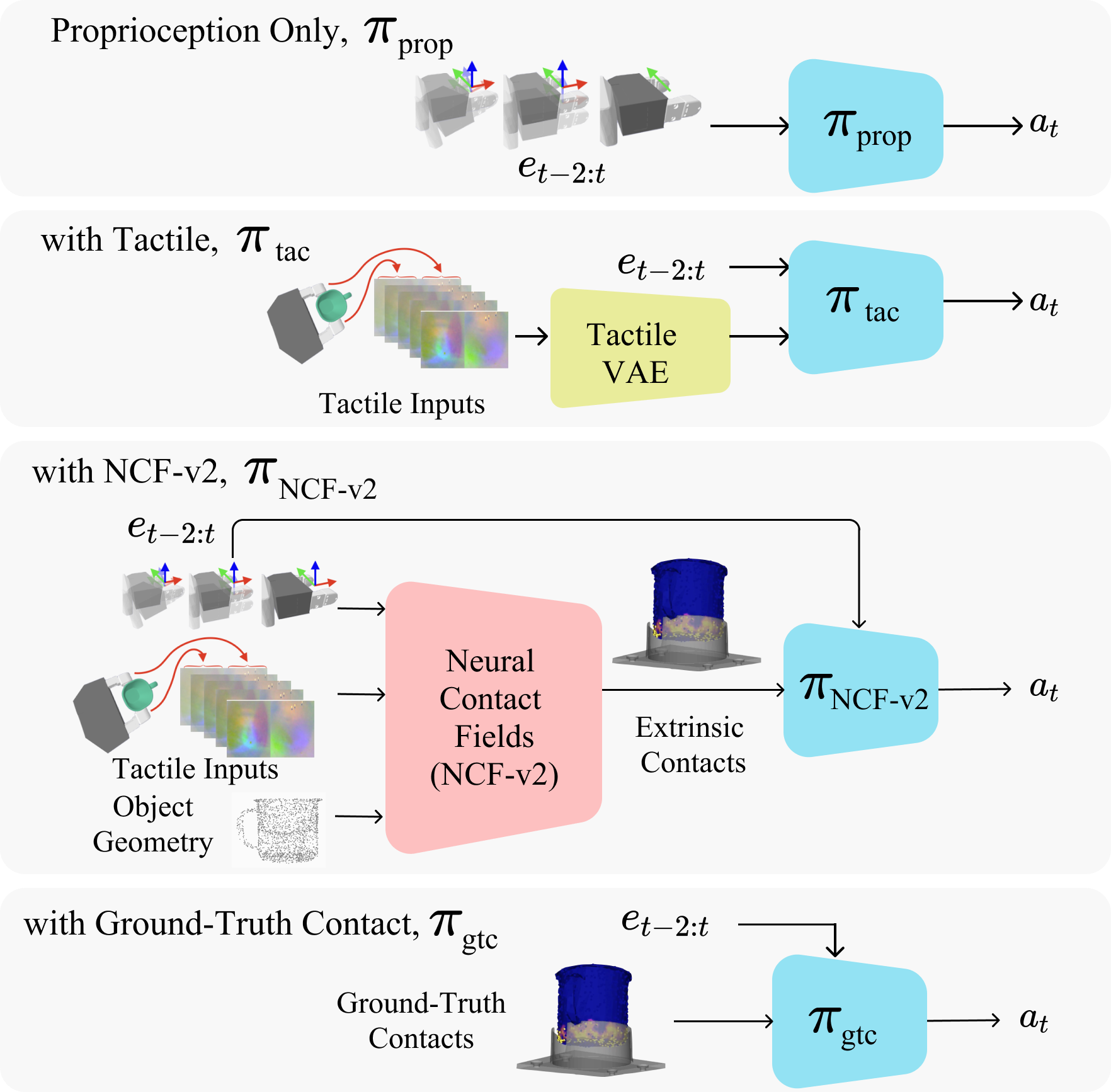}
    \vspace{-5mm}
    \caption{\textbf{Overview of policies trained with different inputs}. We systematically study and compare the utility of estimating extrinsic contact in insertion tasks. We show explicitly estimate extrinsic contacts from tactile sensing is important for successful sim-to-real policies.}
    \label{fig:policies_blocks}
    \vspace{-5mm}
\end{figure}

\textbf{Policy learning.} To have a clear understanding of the utility of estimating extrinsic contacts for policy learning, we train and compare the performance of four policies with different observation space, as illustrated in Fig.~\ref{fig:policies_blocks}.

\begin{itemize}
    \item \textbf{Proprioception Only}, \piprop. This policy only receives end-effector pose information at time steps $t$, $t-1$, and $t-2$. We define these inputs as $\mathbf{e}_{t-2:t} \in \mathbb{R}^{21}$.
    \item \textbf{w/ Tactile, }\pitac. We augment the proprioceptive observations by incorporating the embedding of the sequence of tactile images. This is to test if directly incorporating tactile sensing is sufficient when compared to intermediate estimation of extrinsic contact.
    \item \textbf{w/ \ours, }\pincf: We augment the proprioceptive observations by including the estimated contact patch $\boldsymbol{p}_t$ using \ours. This contact patch corresponds to the points on the object with a contact probability above a threshold. The contact patch is processed through an MLP and its feature embedding is extracted using max-pooling.
    \item \textbf{w/ Ground Truth Contact, }\pigtc: We augment the proprioceptive observations with the point cloud of the ground truth contact patch. It represents an upper bound for evaluation. 
\end{itemize}

For each variant, we jointly optimize the policy $\pi$ and the downsampling MLPs using PPO~\cite{schulman2017proximal}. For the \mug, the reward function encourages the robot to approach the cupholder with the correct orientation, allowing the mug's handle to fit into the cupholder's slot:
$r_t = -\lambda_{dist} \; r_{dist} - \lambda_{rot } \; r_{rot}$,
where $r_{dist}=||\mathbf{k_m}-\mathbf{k_c}||_2$ penalizes the misalignment of four keypoints distributed along the mug central axis ($\mathbf{k_m}$)  and the cupholder axis ($\mathbf{k_c}$). Additionally, $r_{rot}=||\text{ref}_{rot}-\text{mug}_{rot}||_2$ penalizes any deviation in the current orientation of the mug from a reference or goal orientation.
For the \bowl task, the reward function seeks to minimize the misalignment between the position of the bowl $p_{bowl}$ and target slot $p_{target}$:
$r_t = -|| p_{bowl} - p_{target} ||_2$.

\begin{figure}[!t]
    \centering
    \vspace{0.2mm}
    \includegraphics[width=0.47\textwidth]{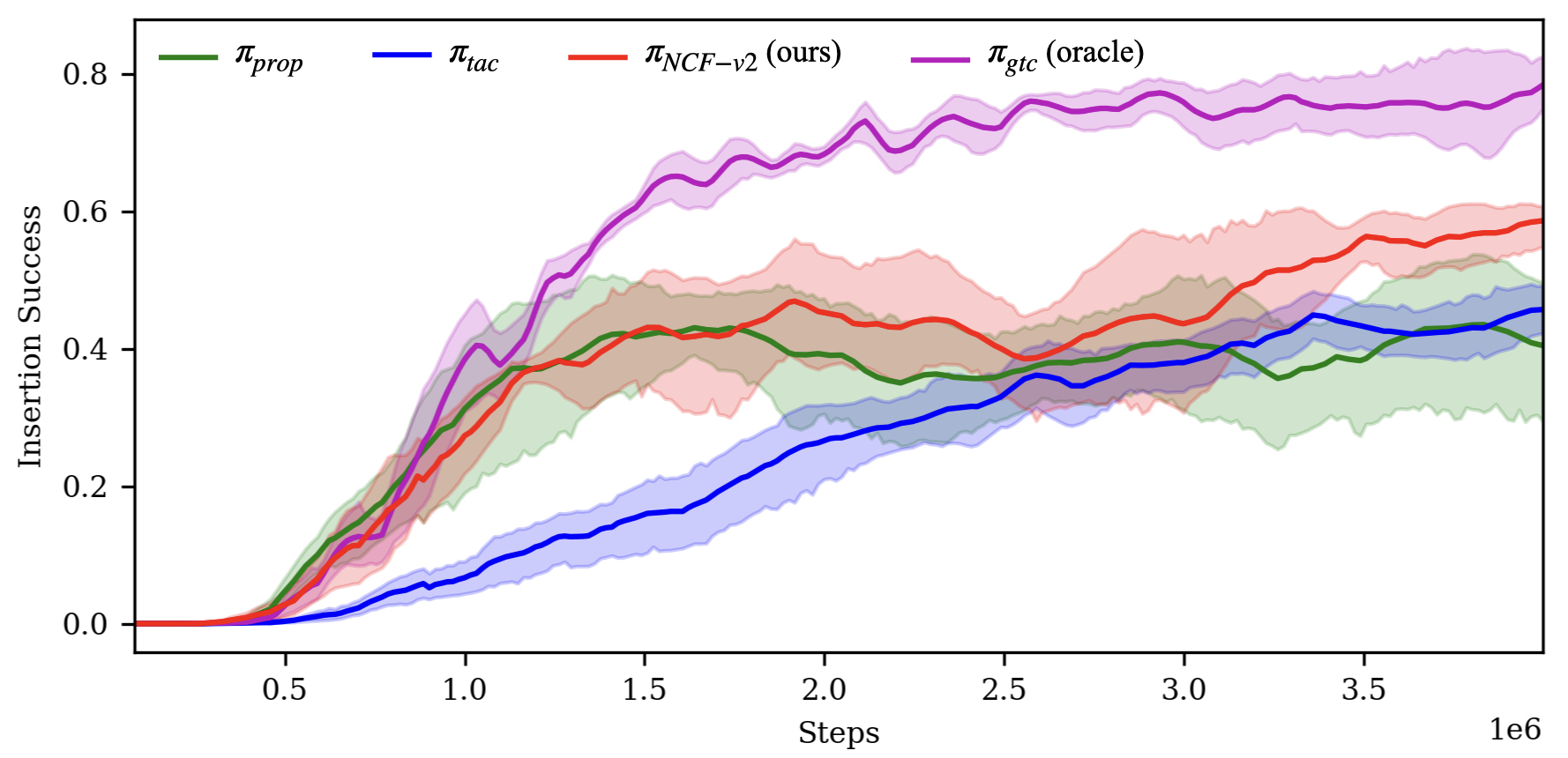}
    \vspace{-3.2mm}
    \caption{Success rate mean and $95\%$ confidence interval for \mug task across 256 parallel simulated environments. Each policy is trained with 5 seeds. \pincf outperforms policies using proprioception only or directly using tactile data.}
    \label{fig:mug_success}
    \vspace{-3mm}
\end{figure}

\begin{figure}[!t]
    \centering
    \includegraphics[width=\linewidth]
    %{figures/mug_cupholder/metrics_cupholder.jpeg}
    {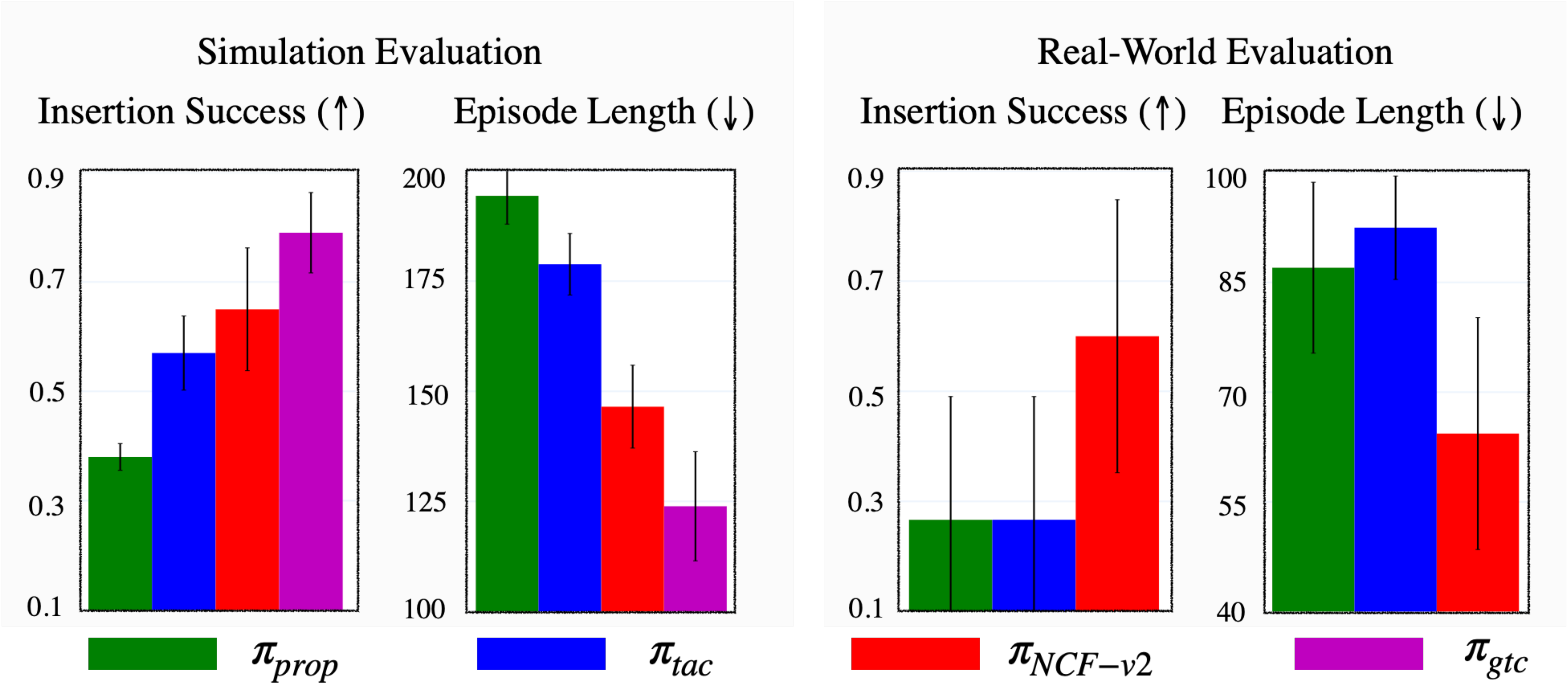}
    \vspace{-6mm}
    \caption{\textbf{Evaluation of mug-in-cupholder policy}. We show insertion success and episode length of 100 trials in simulation (left) and 15 trials in real (right). \pincf performs better and faster compared to other policies that do not leverage extrinsic contact.}
    \label{fig:mug_metrics}
    \vspace{-4mm}
\end{figure}

\subsection{Task 1: \mug}\label{sec:rl_mug_eval}

Learning an insertion policy for this task is inherently challenging due to the narrow tolerance of 1 cm between the test mug and the cupholder. While the robot can reduce keypoint misalignment, successful insertion ultimately hinges on the handle fitting into the slot. Fig.~\ref{fig:mug_success} illustrates the progression of the success rate during policy training. These learning curves strongly affirm the advantage of incorporating extrinsic contacts in learning policies for contact-rich tasks. 

With access to ground truth extrinsic contacts, the policy achieves about $80\%$ success rate in correctly inserting the mug. However, relying solely on the robot's proprioception provides limited information, leading to early convergence to a low-performance policy. While \piprop is good at aligning the objects, it struggles with orienting the mug correctly. Integrating tactile information into the observations proves beneficial, but it demands extensive exploration to effectively map tactile signals to actions. \pitac requires nearly double the number of steps to achieve the same success rate as the proprioception-only policy. In contrast, a policy leveraging extrinsic contacts, like \pincf and \pigtc, demonstrates a relatively straightforward ability to align and orient the mug. This is facilitated by the availability of task-specific information, such as contacts on salient features of the object like the handle, body, and foot. \pincf exhibits a drop in performance but allows to localize extrinsic contacts in real-world. 
% A policy combining proprioception with tactile input is likely to necessitate a considerable number of training episodes.

\begin{figure}
    \centering
    \vspace{0.2mm}
    \includegraphics[width=0.476\textwidth]{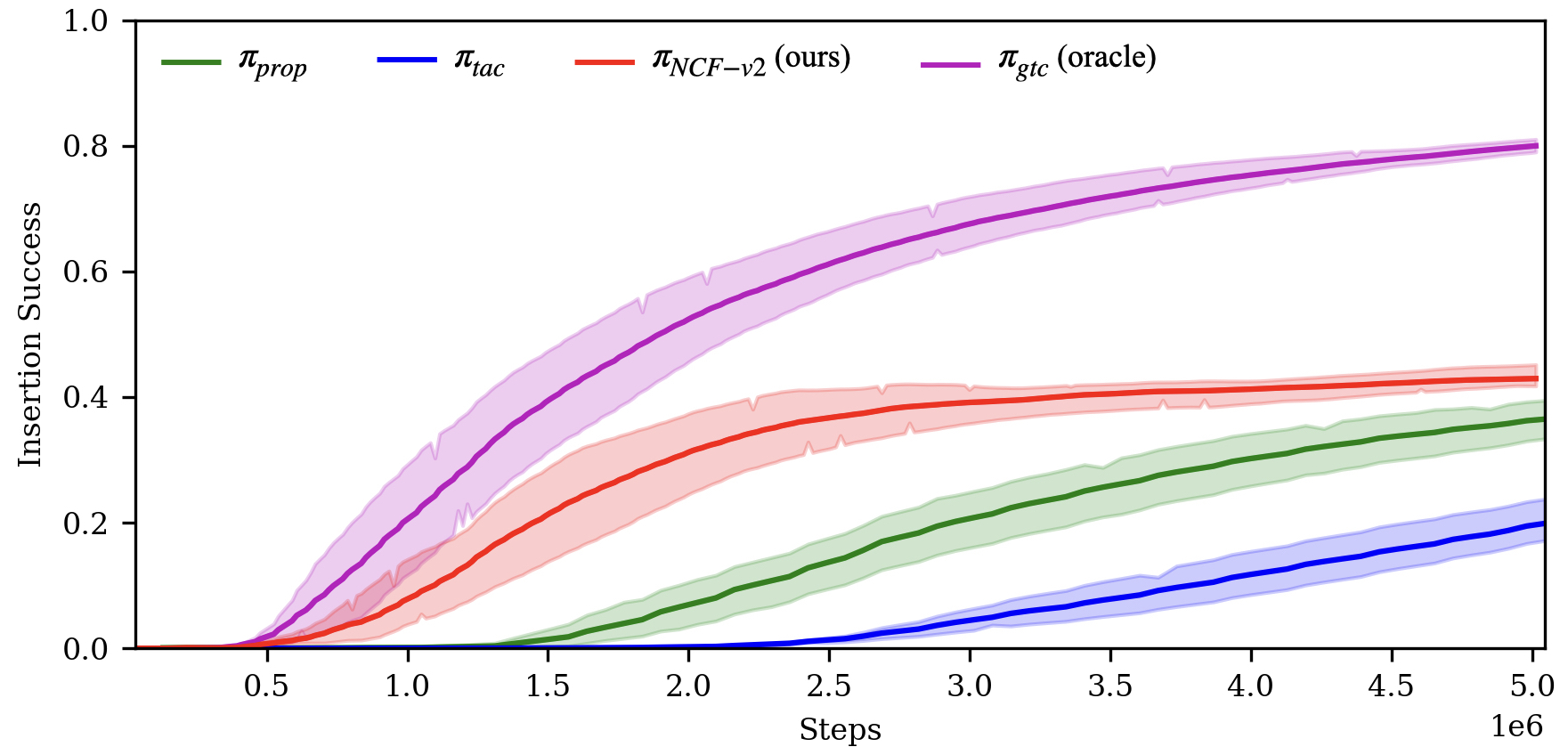}
    \vspace{-3.2mm}
    \caption{Success rate mean and $95\%$ confidence interval for \bowl task across 256 parallel simulated environments. Each policy is trained with 3 seeds. \pincf outperforms policies using proprioception only or directly using tactile data.}
    \label{fig:bowl_success}
    \vspace{-2mm}
\end{figure}

\begin{figure}
    \centering
    \includegraphics[width=0.965\linewidth]{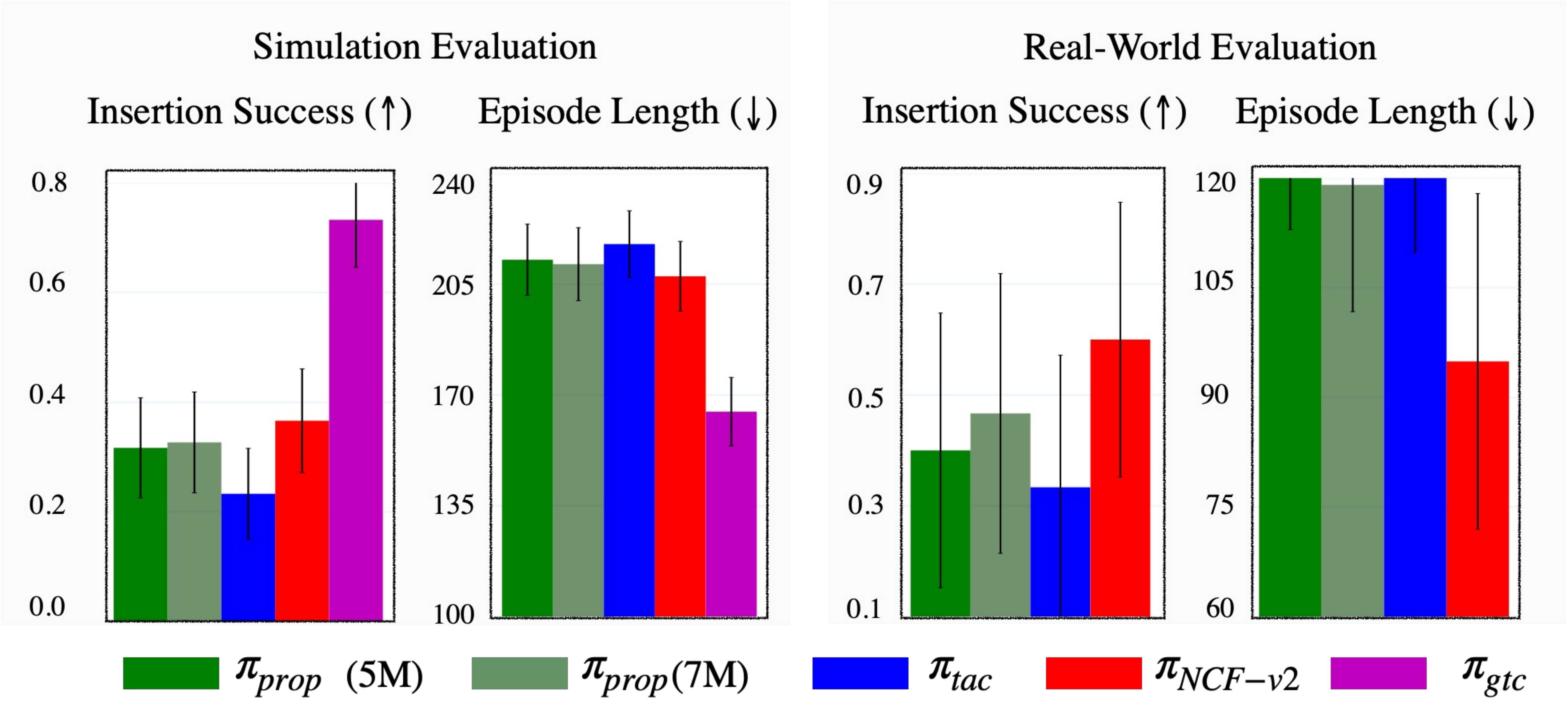}
    \vspace{-1.7mm}
    \caption{\textbf{Evaluation of bowl-in-dishrack policy.} We show success and episode length of 100 trails in simulation (left) and 15 trials in real (right). \pincf performs better and faster compared to other policies that do not leverage extrinsic contact.}
    \label{fig:bowl_metrics}
    \vspace{-4mm}
\end{figure}

We test all policies in simulation, running 100 repetitions. We also perform real-world experiments to evaluate \piprop, \pitac, and \pincf running 15 repetitions, without performing any fine-tuning. For each test, we randomize the initial end-effector pose. Fig.~\ref{fig:mug_metrics} presents the performance metrics for both simulation and real-world experiments. In the simulation experiments, policies that incorporate extrinsic contact information demonstrate the highest success rates. Moreover, leveraging extrinsic contacts provides the additional benefit of learning policies capable of inserting the mug more efficiently in terms of the number of timesteps required to complete the task.

Ultimately, the real-world metrics are the strongest evidence of the utility of extrinsic contacts for the downstream task. \pincf has an average success rate of $60\%$ over 15 repetitions, surpassing \piprop and \pitac at $27\%$. This represents an absolute improvement of $33\%$. Moreover, \pincf proves to be $1.36 \times$ faster, finishing the task on average in 23 steps fewer than its counterparts, with each trial limited to a maximum of 100 steps.

% \pincf consistently outperforms \piprop and \pitac, excelling in both insertion success rate and number of steps required for completing the task. Extrinsic contacts allow the policy to exploit the information about the location of the contact patch, enabling it to rectify undesired collisions (e.g., on the handle) and achieve the intended contacts on the foot and body of the mug. These results mirror our own experiences when attempting an insertion task blindly, relying solely on tactile feedback.

\subsection{Task 2: \bowl}\label{sec:rl_bowl_eval}

This task requires spatial awareness in order to avoid collisions with the dividers when approaching the target slot and to leverage the dividers to guide the insertion. Fig.~\ref{fig:bowl_success} shows the progression of the success rate during policy training. Having observability over extrinsic contacts proves to be very helpful for this task, especially when comparing the progression of the success rate for the oracle policy. Policies with only proprioceptive information or with raw tactile feedback require intensive exploration of the environment. Even with longer training \piprop cannot reach performance comparable to the oracle policy.

For training \pincf, we are using the same model explained in Section~\ref{sec:ncfv2}. This means that we are evaluating the ability of \ours to estimate useful extrinsic contacts for a downstream task under both shape and environment generalization. However, given the small gap between the success rate seen after $5e6$ steps between \piprop and \pincf, we train \piprop for $7e6$ steps to disambiguate the differences in performance when deploying the policies.

We test all policies in simulation, running 100 repetitions. We also perform real-world experiments to evaluate \piprop, \pitac, and \pincf running 15 repetitions, without performing any fine-tuning. For each test, we randomize the initial end-effector pose.  Fig.~\ref{fig:bowl_metrics} presents the performance metrics for both simulation and real-world experiments. In simulation, we find that with privileged contact information the policy correctly inserts the bowl in the desired slot in the dishrack around $75\%$ of the time. However, the other policies exhibit a significantly lower performance. It is worth noting that in simulation we could not completely control the initial pose of the bowl due to undesired collisions with the gripper. Therefore, there was also a high degree of randomization in the initial bowl pose. While the oracle policy could implicitly provide information about the bowl's pose through the contact point cloud, the other policies could not infer the bowl's pose solely from the end-effector information. Even though \ours could estimate the contact pointcloud, it also relies on the end-effector to infer the object's pose.

In the real-world experiments, the initial pose of the bowl is controlled, and all policies exhibit better performance. \pincf has the highest insertion rate, highlighting that \ours is generalizing to an unseen shape and environment. A policy with only proprioceptive information is not able to gather the spatial understanding that this task requires, even if trained for a longer time. \pincf achieves an absolute improvement of $13\%$ in success rate with respect to \piprop and completes the bowl insertion $1.27 \times$ faster.

%%%%%%%%%%%%%%%%%%%%%%%%%%%%%%%%%%%%%%%%%%%%%%%%%%%%%%%%%%%%%%%%%%%%%%%%%%%%%%%
\vspace{-0.5mm}
\section{Conclusion and Limitations}
\vspace{-0.5mm}

In this work, we have demonstrated the utility of estimating extrinsic contacts from touch for learning robot insertion policies. Building upon prior work on \ncf, we first made improvements to the tactile embedding and contact regressor to enable sim-to-real transfer of extrinsic contact estimation during real robot interactions. Through simulation and real-world evaluations, we verified our hypothesis that estimating extrinsic contact aids policy learning for two insertion tasks, \mug and \bowl. Policies with access to estimated/ground-truth extrinsic contacts achieve higher success rates than a baseline policy with access to only proprioception and to raw tactile data. In real-world experiments, policies with \ours outperform the baselines without fine-tuning, achieving $33\%$ higher success and $1.36 \times$ faster execution on \mug, and $13\%$ higher success and $1.27 \times$ faster execution on \bowl.

\textbf{Limitations.}
The current \ncf implementation is restricted to tracking extrinsic contacts for three specific object classes, assuming a fixed relative pose during grasping. Additionally, to get the correct reference point cloud, prior knowledge of the object class is necessary. To overcome these constraints, integrating \ncf into a system that learns an implicit representation of various object shapes through vision can be advantageous.
% Axes for improvement: 
% - Generalization across objects and contact regions
% - Relax requirement for object shape known in advance. 
% Hope to integrate this into a real-time perception system that is used to learn dextrous manipulation policies for contact-rich tasks. 

%%%%%%%%%%%%%%%%%%%%%%%%%%%%%%%%%%%%%%%%%%%%%%%%%%%%%%%%%%%%%%%%%%%%%%%%%%%%%%%
\section*{Acknowledgment}
The authors thank Tess Hellebrekers, Sudharshan Suresh, and Taosha Fan for helpful discussions, Raunaq Bhirangi for help with Franka controller, Mike Lambeta and Roberto Calandra for help with DIGIT sensor, and Elia Rühle for help with TACTO simulator.

%%%%%%%%%%%%%%%%%%%%%%%%%%%%%%%%%%%%%%%%%%%%%%%%%%%%%%%%%%%%%%%%%%%%%%%%%%%%%%%
\clearpage
\balance
\bibliographystyle{IEEEtran}
\bibliography{IEEEabrv, references}

\end{document}